\documentclass[12pt,a4paper]{article}
\usepackage[left=2.2cm,top=3.8cm,bottom=3cm,right=2.2cm]{geometry}
\usepackage[ruled,vlined]{algorithm2e}
\usepackage{dirtytalk}
\usepackage{comment} 
\usepackage{caption}
\usepackage{comment}
\usepackage{amsmath}
\usepackage{float}

\usepackage[utf8x]{inputenc}
\usepackage{amsmath,amssymb}
\usepackage{xcolor}
\usepackage{mathtools}
\usepackage{tabulary}
\usepackage{amsthm}
\usepackage{latexsym}
\usepackage[mathcal]{eucal}
\usepackage{amscd}
\usepackage{graphicx}
\usepackage{amsfonts}
\usepackage{amscd}
\usepackage{MnSymbol}
\usepackage{setspace}
\newtheorem{definition}{Definition}[section]
\usepackage[english]{babel}
\usepackage{listings}
\usepackage{calligra}

\newtheorem{remark}{Remark}[section]
\newtheorem{theorem}{Theorem}[section]
\newtheorem{lemma}{Lemma}[section]
\newtheorem{proposition}{Proposition}[section]
\newtheorem{corollary}{Corollary}[section]
\newtheorem{example}{Example}[section]

\theoremstyle{remark}

\usepackage{graphicx}
\date{}

\newcommand{\bt}{\begin{theorem}}
\newcommand{\et}{\end{theorem}}
\newcommand{\bl}{\begin{lemma}}
\newcommand{\el}{\end{lemma}}
\newcommand{\bexc}{\begin{exercise}}
\newcommand{\eexc}{\end{exercise}}
\newcommand{\bpr}{\begin{proposition}}
\newcommand{\epr}{\end{proposition}}
\newcommand{\bex}{\begin{example}}
\newcommand{\eex}{\end{example}}
\newcommand{\bc}{\begin{corollary}}
\newcommand{\ec}{\end{corollary}}
\newcommand{\bo}{\begin{proof}}
\newcommand{\eo}{\end{proof}}
\newcommand{\bd}{\begin{definition}}
\newcommand{\ed}{\end{definition}}
\newcommand{\br}{\begin{remark}}
\newcommand{\er}{\end{remark}}
\newcommand{\be}{\begin{enumerate}}
\newcommand{\ee}{\end{enumerate}}

\title{
    {\textbf{DeepMediX: A Deep Learning-Driven Resource-Efficient Medical Diagnosis
Across the Spectrum}}\\[0.2cm]
    \author{\large Kishore Babu Nampalle, Pradeep Singh,\vspace{0.1cm} \\ Uppala Vivek Narayan, Balasubramanian Raman \vspace{0.1cm}\\
    \normalsize Department of Computer Science and Engineering \vspace{0.1cm}\\ \normalsize Indian Institute of Technology Roorkee}
}

\date{\today}

\begin{document}

\maketitle

\begin{abstract}
In the rapidly evolving landscape of medical imaging diagnostics, achieving high accuracy while preserving computational efficiency remains a formidable challenge. This work presents \texttt{DeepMediX}, a groundbreaking, resource-efficient model that significantly addresses this challenge. Built on top of the MobileNetV2 architecture, DeepMediX excels in classifying brain MRI scans and skin cancer images, with superior performance demonstrated on both binary and multiclass skin cancer datasets. It provides a solution to labor-intensive manual processes, the need for large datasets, and complexities related to image properties. DeepMediX's design also includes the concept of Federated Learning, enabling a collaborative learning approach without compromising data privacy. This approach allows diverse healthcare institutions to benefit from shared learning experiences without the necessity of direct data access, enhancing the model's predictive power while preserving the privacy and integrity of sensitive patient data. Its low computational footprint makes DeepMediX suitable for deployment on handheld devices, offering potential for real-time diagnostic support. Through rigorous testing on standard datasets, including the ISIC2018 for dermatological research, DeepMediX demonstrates exceptional diagnostic capabilities, matching the performance of existing models on almost all tasks and even outperforming them in some cases.  The findings of this study underline significant implications for the development and deployment of AI-based tools in medical imaging and their integration into point-of-care settings. The source code and models generated would be released at https://github.com/kishorebabun/DeepMediX.
\end{abstract}

\vspace{4mm}

\section{Introduction}
\label{sec:introduction}
Medical imaging is a crucial component of modern healthcare, providing non-invasive insights into the internal structures of the body for the diagnosis and monitoring of numerous diseases \cite{b1}. However, these images must be interpreted, which typically takes time and requires a great deal of experience.
As the volume of medical imaging data continues to burgeon, the demand for efficient, accurate, and automated image interpretation systems is escalating.\\

Deep learning, a recent development in artificial intelligence (AI), has completely changed the medical imaging industry. These methods, in particular Convolutional Neural Networks (CNNs), have achieved accuracy levels comparable to those of medical professionals in a variety of picture identification and classification tasks\cite{b2}. Despite significant advancements in artificial intelligence, there remains a need for a model that not only excels in accuracy but also maintains computational efficiency for scalable deployment. Moreover, current systems often lack a privacy-preserving collaborative learning framework, which is crucial for facilitating knowledge sharing across healthcare institutions while preserving data integrity.\\

\begin{figure}[ht]

    \begin{minipage}[b]{0.2\linewidth}
        \centering
        \centerline{\includegraphics[height=3.6cm, width=3.6cm]{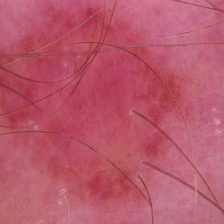}}
        \centerline{Benign}\medskip
    \end{minipage}
    \hfill
    \hspace{-5 mm}
    \begin{minipage}[b]{0.2\linewidth}
        \centering
        \centerline{\includegraphics[height=3.6 cm, width=3.6 cm]{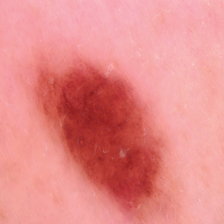}}
        \centerline{Benign}\medskip
    \end{minipage}\hfill
    \hspace{-5 mm}
    \begin{minipage}[b]{.2\linewidth}
        \centering
        \centerline{\includegraphics[height=3.6 cm, width=3.6 cm]{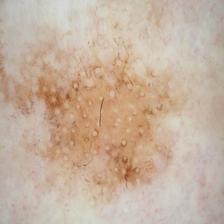}}
        \centerline{Malignant}\medskip
    \end{minipage}\hfill  
    \begin{minipage}[b]{.2\linewidth}
        \centering
        \centerline{\includegraphics[height=3.6 cm, width=3.6 cm]{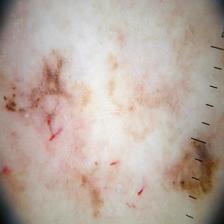}}
        \centerline{Malignant}\medskip
    \end{minipage} \\
    
    \begin{minipage}[b]{.2\linewidth}
        \centering
        \centerline{\includegraphics[height=3.6cm, width=3.6cm]{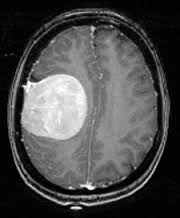}}
        \centerline{Tumor}\medskip
    \end{minipage}
    \hfill
    \begin{minipage}[b]{.2\linewidth}
        \centering
        \centerline{\includegraphics[height=3.6 cm, width=3.6 cm]{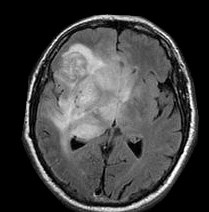}}
        \centerline{Tumor}\medskip
    \end{minipage}\hfill
    \begin{minipage}[b]{.2\linewidth}
        \centering
        \centerline{\includegraphics[height=3.6 cm, width=3.6 cm]{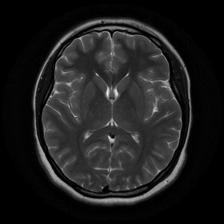}}
        \centerline{No\_Tumor}\medskip
    \end{minipage}\hfill
    \begin{minipage}[b]{.2\linewidth}
        \centering
        \centerline{\includegraphics[height=3.6 cm, width=3.6 cm]{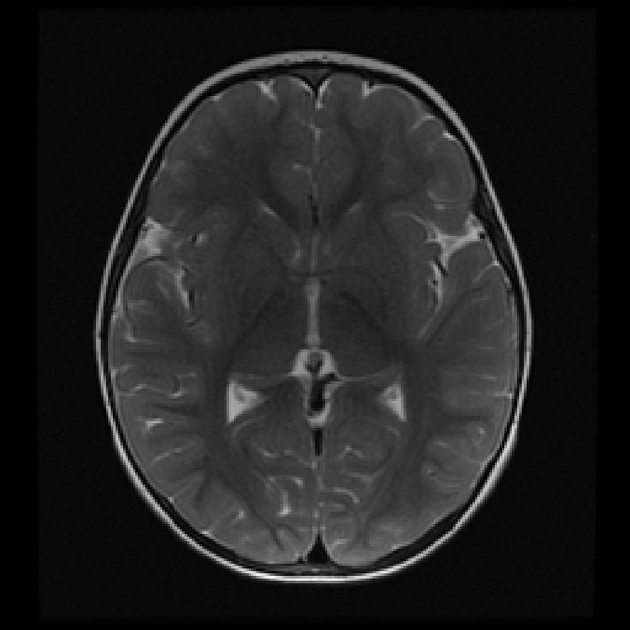}}
        \centerline{No\_Tumor}\medskip
    \end{minipage} \\
    
    \caption{Samples of skin images (Malignant and Benign) and brain images (Tumor and No\_Tumor).}\label{fig:fig1}
    
\end{figure}

This work proposes a novel approach for medical image classification using a refined deep-learning model, focusing
on optimizing performance while maintaining a low computational footprint that can be deployed at scale. The model is built on the MobileNetV2 architecture and augmented with strategic modifications to enhance its performance on specific medical imaging datasets:
brain MRI scans and skin cancer images. Few samples are shown in figure \ref{fig:fig1}. These datasets were selected for their distinct challenges, providing a comprehensive benchmark for the model’s performance. A significant aspect of this research is the development of a model that not
only achieves superior accuracy but also remains computationally efficient. The lightweight nature of the model ensures
its suitability for deployment on handheld devices, making advanced diagnostic support accessible in resource-limited
settings. Furthermore, the model’s design allows for easy scalability, facilitating its adaptation to a growing number of
imaging modalities and larger datasets. The paper's primary contributions are :
\vspace{0.5mm}
\begin{itemize}
    \item We propose a robust, efficient, and scalable model, named DeepMediX for medical image classification.
    
    \item We incorporate Federated Learning into the model, enabling privacy-preserving collaborative learning across healthcare institutions, thus enhancing model performance while maintaining data integrity.

    \item  We explore the model’s performance in diverse medical imaging scenarios, contributing valuable insights to the ongoing discourse on the applicability of deep learning techniques in medical imaging.
\end{itemize}
\vspace{3mm}

The remaining part of the paper is organized as follows: Section 2 includes the literature survey; Section 3 consists of the methodology, including data preparation, model architecture, and training; Section 4 presents the results and an ablation study; Sections 5 and 6 provide a discussion and conclusion of the paper with implications and suggestions for further research respectively.


\section{Background}
The implications of inaccurate conclusions in areas essential to safety can be severe, posing a serious threat to human
life, despite the fact that this particular faulty correlation is only one of many uncovered by careful data analysis. It is
appropriate for medical experts to be wary of relying on the diagnostic projections produced by such systems due to the
limited transparency surrounding how they arrive at their judgments \cite{b3}. Even though Deep Learning (DL) techniques
for Computer-Aided Diagnosis (CAD) have been widely used in the recent decade \cite{b4}\cite{b5}, this reluctance still exists.\\

Recently, many AI-based systems for classifying medical images have been proposed \cite{b6}. However, some frameworks
don’t provide adequate explanations that are easy to grasp, while others don’t provide a user-friendly interface for
human-machine interaction. These flaws prevent these frameworks from being effectively used in diagnostic procedures
or research. Additionally, the first commercial platforms for biological AI have started to appear \cite{b7}, and they claim to
be able to explain their methods. The integration of AI in healthcare, especially in medical imaging, has been a subject of
intense research over the past decade. The amalgamation of advanced AI techniques, particularly deep learning, with the
intricate realm of medical imaging has unveiled exciting possibilities in diagnostics and treatment planning \cite{b8}. Different
techniques for presenting and improving these ideas lead to varied levels of understanding of how AI makes decisions.
AI image-based classifiers can use a variety of methods, each using a different strategy to provide explanations. These
methods involve visualizing \cite{b9} \cite{b10} the relevance of characteristics, providing textual descriptions \cite{b11}\cite{b12}, or using
quantitative methods to gauge the value of abstract ideas \cite{b13}.
In particular, CNN has been used extensively in research studies to analyze dermoscopic images and identify both common and malignant skin diseases \cite{b14}\cite{b15}. Since dermoscopic images often need
dermatoscopes, frequently available in dermatology clinics, other studies have concentrated on using deep learning
techniques with clinical manifestations to identify skin problems \cite{b16}\cite{b17}. According to numerous works \cite{b18}\cite{b19},
deep learning–based systems can identify illnesses using dermoscopic and clinical images with similar accuracy as skilled
dermatologists. CNNs, a class of deep learning models, have been at the forefront of numerous innovations in medical
image analysis. These models excel at identifying patterns in images and learning complex representations without the
need for manual feature extraction. Deep learning-based algorithms have shown high efficacy in various image-based
tasks such as detection, classification, segmentation, and even in more complex tasks like predicting patient prognosis \cite{b20}.
Medical imaging is typically carried out manually by knowledgeable specialists like radiologists, sinologists, or pathologists. Medical imaging methods \cite{b21} such as computed tomography (CT), magnetic resonance imaging (MRI), ultrasound imaging, and histopathological imaging are often employed and are very effective at recognizing objects when compared to other testing methods.\\


In the realm of magnetic resonance imaging (MRI), AI has made significant
strides, especially in the interpretation of brain MRIs. Techniques like 3D CNNs and U-Nets have been employed
for the segmentation of brain structures and for the detection of abnormalities such as tumors \cite{b22}. Moreover, deep
learning has shown potential in the early detection of neurodegenerative diseases like Alzheimer’s from brain MRIs \cite{b24}.
Although traditional techniques are efficacious, they frequently rely on medical data with a standard grid arrangement, which may curb their complete potential. A novel methodology has been introduced that converts this grid-centric data into more intricate representations using point clouds and surface delineations. This subsequently facilitates a more thorough image examination \cite{b25}.
Recently, CNN-based models have been deployed for medical image classification. Among many instances, a CNN-
based classifier outperformed 21 board-certified dermatologists in successfully classifying clinical and dermoscopic
images \cite{b26}. Based on the review of existing studies, there currently needs to be more information regarding the utilization
of artificial and human intelligence together in the domain of medical computer vision. Specifically, previous research has
primarily focused on comparing the performance of humans and machines in the area of skin cancer detection. Skin
cancer detection is another domain where AI has shown impressive results. CNNs have been used to classify skin lesions,
showing comparable performance to dermatologists in differentiating malignant melanomas from benign nevi \cite{b27}\cite{b28}.
Despite these advances, there is still a vast scope for improvement. A major challenge lies in building models that
are not only accurate but also computationally efficient. Many deep learning models have large computational demands,
making them less suitable for use in real-time applications or deployment on handheld devices. MobileNetV2 is a model
specifically designed to address this issue, providing high accuracy with a significantly reduced number of parameters and
thus computational requirements, compared to other models \cite{b29}. This work builds upon these developments, adapting
and enhancing the MobileNetV2 architecture for efficient medical image classification in diverse imaging scenarios. The
model’s performance is evaluated on brain MRI and skin cancer image datasets, benchmarking its effectiveness in these
specific medical contexts.

\section{Methodology}

This section details the overall methodology employed in this study, which includes data preparation, architecture of the deep learning model, and the training process.

\subsection{Mathematical Foundations of the Model}
CNNs use convolution operations, mathematically represented as: $F(i, j) = \sum_m \sum_n I(i - m, j - n)K(m, n)$ where $F$ is the feature map, $I$ the image, and $K$ the kernel. In particular, MobileNetV2 applies depthwise separable convolutions, two-layer operations that combine depthwise and pointwise convolutions. In terms of computational efficiency, these surpass conventional convolutions and can be calculated as follows: \(Y_{kj} = \sum_{l=0}^{M-1} K_{kl} \cdot X_{lj}\). In this equation, \(X\) denotes the input, \(K\) symbolizes the filter, \(Y\) indicates the output, and the \say{\(\cdot\)} operator signifies the dot product. CNNs use shared weights to achieve translation invariance, recognizing features regardless of their location. Other invariances such as rotation or scale are facilitated through data augmentation. Feature maps output by CNNs visualize high activations corresponding to detected features. The ReLU function is a commonly used activation function in deep learning, defined as $f(x) = \max(0, x)$. It promotes computational efficiency and sparsity in the network, activating only a subset of neurons at a given time. Dropout mitigates overfitting by randomly \say{dropping out} layer outputs during training with a given probability $p$. It can be represented for a single neuron as $y_i = r_i x_i$ where $x_i$ and $y_i$ are input and output neurons respectively, and $r_i$ is a Bernoulli random variable with probability $p$. Global Average Pooling (GAP) replaces fully connected layers to reduce overfitting by reducing parameter count. GAP averages all values in each feature map, calculated as $y_c = \frac{1}{H \times W} \sum_{i=1}^{H} \sum_{j=1}^{W} f_{ij}$ where $f_{ij}$ is the feature map value at location $(i, j)$ and $y_c$ is the c-th output element of the GAP layer. The Adam optimizer extends stochastic gradient descent. The update rule for each weight is defined as:
$m_t = \beta_1 m_{t-1} + (1 - \beta_1) g_t$, $v_t = \beta_2 v_{t-1} + (1 - \beta_2) g_t^2$, $\hat{m}_t = \frac{m_t}{1 - \beta_1^t}$, $\hat{v}_t = \frac{v_t}{1 - \beta_2^t}$, and $\theta_t = \theta_{t-1} - \alpha \frac{\hat{m}_t}{\sqrt{\hat{v}_t} + \epsilon}$, where $\theta_t$ represents the parameters of the model, and $g_t$ the gradient at time step $t$. The binary cross-entropy loss, defined as $L(y, \hat{y}) = -\frac{1}{N} \sum_{i=1}^{N} [y_i \log(\hat{y}_i) + (1 - y_i) \log(1 - \hat{y}_i)]$, is used for binary classification. For multi-class classification, we use the categorical cross-entropy loss: $L(y, \hat{y}) = -\frac{1}{N} \sum_{i=1}^{N} \sum_{c=1}^{C} y_{ic} \log(\hat{y}_{ic})$. These functions penalize probabilities far from the true labels.\\

The Universal Approximation Theorem asserts that a neural network with a single hidden layer can approximate any continuous function, given certain conditions. Consider a function $\phi: \mathbb{R} \to \mathbb{R}$ that is neither constant nor unbounded, and increases monotonically while remaining continuous. For any function $f$ from the class $C(I_m)$ and any positive $\epsilon$, we can find constants $w_{i,j}, b_i, \theta_i$ that allow the approximation $F(x_1,...,x_m) = \sum_{i=1}^{N} \theta_i \phi\left(\sum_{j=1}^{m} w_{i,j} x_j + b_i\right)$ to hold true, where the difference between $F(x)$ and $f(x)$ is less than $\epsilon$ for all $x$ in $I_m$. In the context of medical imaging and the specific network designed in this study, the Universal Approximation Theorem has significant implications. The theorem underpins the network's ability to approximate any continuous function - in this case, the mapping from the input image data to the output classes (brain MRI or skin cancer status). Given a suitable number of neurons and an appropriate choice of activation function, our network can, theoretically, learn to approximate the true underlying distribution of the data. Let's denote our input data (medical images) as $I \in \mathbb{R}^{224 \times 224 \times 3}$ and our output as $y \in \{0,1\}$ representing the class labels (normal or abnormal). For simplicity, let's represent our neural network function as $F_\theta(I)$, where $\theta$ represents the parameters (weights and biases) of the model. Our objective during training is to find the optimal $\theta^*$ such that $F_{\theta^*}(I)$ is a good approximation of the true labels $y$. As per the Universal Approximation Theorem, for any given $\epsilon > 0$, a $\theta^*$ can be found such that for every input image $I$, it satisfies: the average of absolute difference between $F_{\theta^*}(I)$ and $y$ is less than $\epsilon$. This suggests that our network has the potential to learn this complex function mapping, provided we have enough computational resources and data and assume that the function is within the scope of what a neural network can represent. However, it's important to note that while the theorem assures us that such a $\theta^*$ exists, it doesn't provide any guidance on how to find it. This is where the importance of a well-designed architecture, a suitable optimizer, and an appropriate loss function comes into play. These tools guide the learning process, helping us find a good $\theta^*$ that can accurately map the input images to their correct labels.\\

\subsection{Federated Learning}

 Federated learning is a distributed variant of machine learning wherein a model is trained on numerous devices or servers that each store local data samples, all without any data transfer. This strategy enables training on a vast body of data that is distributed across devices, like mobile phones, proving particularly beneficial in sectors where privacy is of utmost importance, such as healthcare. Despite its benefits, federated learning presents challenges concerning privacy and security as clients' data should not be exposed to the server or other clients. Techniques such as differential privacy and secure multi-party computation can mitigate these risks.\\

\begin{algorithm}[ht]
\SetAlgoLined
\textbf{Input:}
\begin{enumerate}
    \item An initial global model $w_0$.
    \item A collection of $K$ clients, each with their own local dataset $D_k$.
    \item Local update method (either \say{SGD} or \say{SVRG}).
\end{enumerate}

\textbf{Output}:  A trained global model $w^*$.\\

\textbf{Algorithm}:

\begin{enumerate}
    \item \textbf{Initialization}: Let $t=0$ and set the global model $w_t$ equal to the initial global model $w_0$.
    \item \textbf{For} a number of communication rounds \textbf{do}:
    \begin{enumerate}
        
    \item  \textbf{Broadcast}: The server sends the current global model $w_t$ to a selected subset of clients.
    \item \textbf{Local Update}: Each client $k$ updates their\\ model to $w_{k,t}$ based on its local data $D_k$ and\\ the chosen update method.
    \begin{enumerate}
        \item If \say{SGD}, then $w_{k,t} = \text{SGD}(w_t, D_k)$.
        \item If \say{SVRG}, then each client $k$ computes the full gradient of its local loss function at the current model $w_t$:
        $$g_{k,t} = \nabla F_k(w_t)$$
        and performs several steps of SVRG, using\\ its local dataset $D_k$ and the full gradient\\ $g_{k,t}$. This results in an updated model $w_{k,t}$.
    \end{enumerate}
    \item \textbf{Aggregate}: The server collects the updated models $w_{k,t}$ from each client and aggregates them to update the global model. This can be done by computing a weighted average: 
    $$w_{t+1} = \frac{\sum_{k=1}^{K} n_k \cdot w_{k,t}}{\sum_{k=1}^{K} n_k}$$
    where $n_k$ denotes the samples in client $k$'s dataset $D_k$.
    \item \textbf{Increment} $t$.
    \end{enumerate}
    \item \textbf{Return} the final global model $w^* = w_t$ as the output.
\end{enumerate}
\caption{Generalized Federated Learning Algorithm}
\label{alg:algo1}
\end{algorithm}

In the federated learning process as shown in algorithm \ref{alg:algo1}, each client, holding a local dataset $D_k$ of size $n_k$, computes an updated model $w_{k,t}$ based on its own local data given a global model $w_t$ at time $t$. This update is typically achieved by running several epochs of an optimization algorithm, such as Stochastic Gradient Descent (SGD) or Stochastic Variance Reduced Gradient (SVRG) on the local data. The SVRG method, which involves computing the full gradient of the local loss function at the current model and using this full gradient in the update rule, reduces the variance of the gradient estimates, potentially leading to faster convergence. The server collects the updated models $w_{k,t}$ from each client and aggregates them to update the global model. A typical method for this is computing a weighted average, where the weights are proportional to the number of data samples on each client. The objective of federated learning is to minimise the global loss, equivalent to minimizing the expected loss over the entire data distribution, assuming each client's data is an independently and identically distributed (i.i.d.) sample from this distribution. This assumption, however, does not always hold in practice due to different clients possibly having very different types of data (non-IID data). Solutions to this issue may involve more sophisticated aggregation methods or adjustments to the local update procedure.\\

\subsection{Data Preparation}

Both the brain MRI and skin cancer datasets were subjected to preprocessing steps to ensure that they were suitably prepared for the model. This included resizing the images to fit the input shape of the model (\texttt{224, 224, 3}), and normalizing the pixel values to be within the range of 0-1. 
\subsection{Model Architecture}

The architecture of MobileNetV2, which serves as the backbone of DeepMediX, is characterized by the use of inverted residuals and linear bottlenecks. Mathematically, this design involves the use of an expand layer, a depth-wise convolutional layer, and a projection layer. The expand layer is defined as $H^{'} = H_1 * (1 + t)$, where \(H^{'}\) is the expanded dimension, \(H_1\) is the input dimension, and \(t\) is the expansion factor, typically set to 6 in MobileNetV2. The depth-wise convolution is a space-wise application of convolution operation, which reduces the computational cost and model parameters by independently applying convolution filters to each input channel: $H^{''} = H^{'} * K * K * M$, where \(H^{''}\) is the output of the depth-wise convolution, \(K\) is the kernel size, and \(M\) is the number of input channels. The projection layer uses a \(1\times 1\) convolution to project the feature map back to a lower dimensional space: $H^{'''} = H^{''} * 1 * 1 * N$, where \(H^{'''}\) is the final output, and \(N\) is the number of output channels. The use of a linear activation function in this layer, instead of a non-linear function like ReLU, is to prevent loss of information from the high-dimensional space. This model, pretrained on ImageNet \cite{b43}, is known for its efficiency and performance on a variety of vision tasks. We constructed the top over the base for the classification task as follows.\\

Let $\mathcal{X}$ represent the space of medical images. Our aim is to define a mapping $f: \mathcal{X} \rightarrow \mathcal{Y}$, where $\mathcal{Y}$ is the set of medical conditions or labels.  First, the model processes an image $x \in \mathcal{X}$ using the pretrained MobileNetV2 architecture. Let $f_{\text{base}}: \mathcal{X} \rightarrow \mathbb{R}^{d}$ represent this pretrained MobileNetV2 function, where $d$ is the dimensionality of the output feature vector. Hence, we get the representation of an image as
$z = f_{\text{base}}(x)$, where $z \in \mathbb{R}^{d}$ is the output feature vector. Next, we add several additional layers to the network. Let $f_{\text{add}}: \mathbb{R}^{d} \rightarrow \mathbb{R}^{k}$ represent these added layers, where $k$ is the number of classes. Thus, we get $y' = f_{\text{add}}(z)$, where $y' \in \mathbb{R}^{k}$ is the raw output of the network. Then, we apply a softmax function to the output of these additional layers to obtain the final class probabilities. Let $\sigma: \mathbb{R}^{k} \rightarrow \mathbb{R}^{k}$ represent the softmax function, such that $\sigma(y')_i = \frac{e^{y'_i}}{\sum_{j=1}^{k}e^{y'_j}} \quad \text{for } i = 1, \ldots, k$.\\

Ergo, $y = \sigma(y')$, where $y \in \mathbb{R}^{k}$ is the final output of the network, and each element $y_i$ can be interpreted as the probability of the input image $x$ belonging to class $i$. Therefore, the function $f$ for the entire network can be written as a composition of the above functions as $f = \sigma \circ f_{\text{add}} \circ f_{\text{base}}$. For learning, the model uses a standard cross-entropy loss for training. Given the ground truth label $c \in \{1, \ldots, k\}$ and the predicted probabilities $y = (y_1, \ldots, y_k)$, the cross-entropy loss $L$ is defined as $L(y, c) = -\log(y_c)$. The parameters of the added layers are learned by minimizing this loss over the training set. $f_{\text{add}}$, added to the MobileNetV2 base model is defined as follows. Let's consider the initial tensor representation $z \in \mathbb{R}^{h \times w \times d}$ after the MobileNetV2 base model, where $h$ and $w$ are the height and width of the feature map, and $d$ is the number of channels. We then apply several operations:

\begin{itemize}
\item  Dropout layer ($f_{\text{drop1}}: \mathbb{R}^{h \times w \times d} \rightarrow \mathbb{R}^{h \times w \times d}$) with rate 0.4.
\item  Global average pooling layer ($f_{\text{gap}}: \mathbb{R}^{h \times w \times d} \rightarrow \mathbb{R}^{d}$).
\item  Flattening layer ($f_{\text{flat}}: \mathbb{R}^{d} \rightarrow \mathbb{R}^{d}$) where the dimensionality remains the same.
\item  Dropout layer ($f_{\text{drop2}}: \mathbb{R}^{d} \rightarrow \mathbb{R}^{d}$) with rate 0.4.
\item  Dense layer ($f_{\text{dense1}}: \mathbb{R}^{d} \rightarrow \mathbb{R}^{64}$) with 64 neurons and ReLU activation.
\item  Batch Normalization layer ($f_{\text{bn1}}: \mathbb{R}^{64} \rightarrow \mathbb{R}^{64}$).
\item  Dropout layer ($f_{\text{drop3}}: \mathbb{R}^{64} \rightarrow \mathbb{R}^{64}$) with rate 0.4.
\item  Dense layer ($f_{\text{dense2}}: \mathbb{R}^{64} \rightarrow \mathbb{R}^{32}$) with 32 neurons and ReLU activation.
\item  Batch Normalization layer ($f_{\text{bn2}}: \mathbb{R}^{32} \rightarrow \mathbb{R}^{32}$).
\end{itemize}
\vspace{3mm}
We have another path in parallel to the above steps, directly connecting output of $f_{\text{flat}}$ and $f_{\text{bn2}}$. We then concatenate these two paths, which can be expressed as $f_{\text{concat}}: \mathbb{R}^{d} \times \mathbb{R}^{32} \rightarrow \mathbb{R}^{(d + 32)}$. After concatenation, the following operations are applied:
\begin{itemize}
\item  Dropout layer ($f_{\text{drop4}}: \mathbb{R}^{(d + 32)} \rightarrow \mathbb{R}^{(d + 32)}$) with rate 0.4.
\item  Dense layer ($f_{\text{dense3}}: \mathbb{R}^{(d + 32)} \rightarrow \mathbb{R}^{16}$) with 16 neurons and ReLU activation.
\item  Batch Normalization layer ($f_{\text{bn3}}: \mathbb{R}^{16} \rightarrow \mathbb{R}^{16}$).
\item  Dropout layer ($f_{\text{drop5}}: \mathbb{R}^{16} \rightarrow \mathbb{R}^{16}$) with rate 0.4.
\item  Dense layer ($f_{\text{dense4}}: \mathbb{R}^{16} \rightarrow \mathbb{R}^{k}$) with $k$ neurons (4 in this case) and softmax activation, providing the final output $y' \in \mathbb{R}^{k}$.
\end{itemize}

Hence, $f_{\text{add}}$ is defined as:
\begin{align*}
f_{\text{add}} = &f_{\text{dense4}} \circ f_{\text{drop5}} \circ f_{\text{bn3}} \circ f_{\text{dense3}} \circ f_{\text{drop4}} \circ f_{\text{concat}} \\
&\circ \left( f_{\text{bn2}} \circ f_{\text{dense2}} \circ f_{\text{drop3}} \circ f_{\text{bn1}} \circ f_{\text{dense1}} \circ f_{\text{drop2}} \right. \\
&\left. \circ f_{\text{flat}} \circ f_{\text{gap}} \circ f_{\text{drop1}}, f_{\text{flat}} \right)
\end{align*}

\begin{figure}[H]
\centering
\captionsetup{justification=centering}
\includegraphics[width=.9\textwidth]{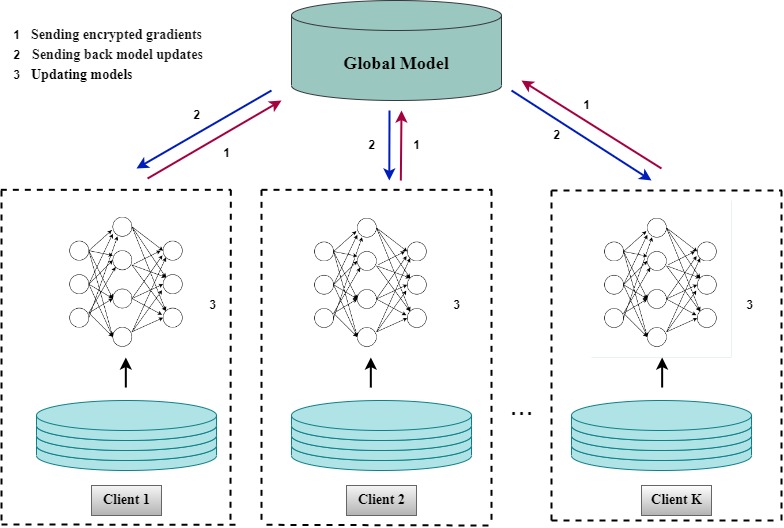}
\vspace{7pt}
\caption{Schematic representation of Federated Learning } \label{fig2}
\end{figure}

As we are working with sensitive medical data in our project, we decided to incorporate federated learning into our DeepMediX model. We used TensorFlow Federated (TFF), an open-source framework for machine learning and other computations on decentralised data, to incorporate federated learning.\\

The overall process involves initializing a global model on a server and then conducting multiple rounds of training, where in each round, the server broadcasts the global model to selected clients (devices with local data), the clients update their models based on their local data, and then send these updates back to the server. The server then aggregates these updates to improve the global model. This procedure persists in a repeated manner until a certain termination condition has been satisfied. Federated Learning is aggregated across multiple \say{clients} (in this case, healthcare institutions) without sharing raw data between them. This is performed by several rounds of computation on each client, followed by aggregation at the central server. We'll represent $K$ as the aggregate count of clients, $n_k$ as the quantity of data samples at the $k$-th client, and $N = \sum_{k=1}^{K} n_k$ as the sum total of data samples spanning all clients. In each round of federated learning as shown in figure \ref{fig2}, the following steps are performed:
\begin{itemize}
\item  \textbf{Sending the current global model to clients}: Each client $k$ receives the current global model parameters, denoted by $w$.

\item  \textbf{Local model training on each client}: Each client $k$ updates the model parameters by applying $E$ steps of SGD (or another optimizer) on its local data. The SGD learning rate is 0.1 as per our code. Let $w_k^{(i)}$ denote the parameters of the model at the $i$-th step of local training on client $k$. The local training can be mathematically represented as $w_k^{(i+1)} = w_k^{(i)} - \eta_k \cdot \nabla L(w_k^{(i)}; x, c)$, for $i=0,\ldots,E-1$, where $(x, c)$ is a mini-batch of local data, $\eta_k$ is the local learning rate and $L$ is the cross-entropy loss function. At the end of the $E$ steps, each client obtains the final local model parameters $w_k = w_k^{(E)}$.

\item  \textbf{Sending local model updates to the server}: Each client sends its local model update $\Delta w_k = w_k - w$ to the server.

\item  \textbf{Aggregating local model updates on the server}: The server computes a weighted average of the local model updates as $\Delta w = \frac{1}{N} \sum_{k=1}^{K} n_k \cdot \Delta w_k$. This represents the average model update across all clients, weighted by the number of data samples at each client.

\item  \textbf{Updating the global model on the server}: The server updates the global model parameters as $ w = w + \Delta w$.

\item  \textbf{Repeat}: The process is repeated for a number of rounds.
\end{itemize}
\vspace{3mm}
We  implemented  federated learning process  using `tensorflow\_federated`. The `state` variable holds the state of the federated learning process, including the global model parameters. The `iterative\_process.next()` function performs one round of federated learning and updates the `state`. The advantage of this approach is that it allows us to train the DeepMediX model on a large amount of decentralized data, which could potentially lead to improved model performance. At the same time, it ensures that the individual data samples do not leave the clients, thus preserving data privacy.

\subsection{Geometrical Interpretation of DeepMediX}
A manifold can be thought of as a space that locally resembles Euclidean space, meaning that while the entire space might be curved, any small region looks flat. In the context of machine learning, a manifold is often used to refer to the structure that data points form in high-dimensional space. An interesting aspect of many machine learning problems, particularly in image and signal processing, is that they naturally reside on a manifold in a high-dimensional space. This is the basis for manifold learning, which posits that while data may live in a very high-dimensional space, it actually occupies a much smaller subspace: a manifold of much lower dimension. This is a particular instance of the manifold hypothesis, which contends that low-dimensional manifolds enmeshed in high-dimensional space contain real-world high-dimensional data.\\


In particular, let's consider how the loss landscape of a neural network can be described in terms of Riemannian geometry \cite{b44}, which studies smooth manifolds equipped with a metric tensor that allows for the measurement of lengths and angles. Consider a deep learning model parametrized by a set of parameters $\theta \in \mathbb{R}^p$. The loss function $L: \mathbb{R}^p \rightarrow \mathbb{R}$ associates a loss value with each set of parameters, thereby defining a loss landscape over the parameter space. By viewing the parameter space as a Riemannian manifold, the gradient of the loss function can be interpreted as a vector in the tangent space of the manifold at the current point $\theta$. This gives us a way to define the concept of \say{direction} in the parameter space, which is fundamental to the operation of gradient-based optimization methods. The Hessian of the loss function, which is a second-order derivative, can be viewed as a type of metric tensor on the parameter space. This allows us to measure distances and angles between different directions in the tangent space, and thus to define the concept of \say{curvature} of the loss landscape.\\

The curvature of the loss landscape is important for understanding the optimization dynamics of neural networks. For instance, regions of high curvature correspond to \say{narrow valleys} in the loss landscape, which are difficult for gradient-based methods to traverse. This can lead to slow convergence or the model getting stuck in suboptimal solutions. On the other hand, regions of low curvature correspond to \say{flat valleys}, which are easier to traverse and can lead to more robust solutions that generalize better. Thus, by studying the geometry of the loss landscape, we can gain insights into the behavior of deep learning models.\\

The process of training a deep learning model can be viewed as an instance of learning a manifold. The neural network effectively learns to transform the data manifold from a complex, convoluted shape in the input space to a more simplified shape in the output space. The task of the machine learning algorithm is to represent the manifold that each data point in a finite dataset occupies in a high-dimensional space. The model learns the manifold structure more precisely with each iteration of the learning process. In terms of federated learning, each client can be viewed as learning a portion of the overall data manifold based on its local data. When these local models are aggregated on the server, the global model becomes a better representation of the entire data manifold.\\


Each layer of the neural network performs a transformation of the data manifold, and these transformations can be described by a set of smooth functions. The key point here is the concept of a Jacobian matrix, denoted by $J(x)$, which can be seen as a linear approximation of the function at a given point $x$, or more precisely, as the best linear approximation in a neighborhood of that point. The Jacobian provides a linear mapping from the input tangent space to the output tangent space, which allows us to study how the network transforms the data manifold locally. By further defining a Riemannian metric $G$ on the data manifold, we can construct a metric on the feature space using the Jacobian. More specifically, given an input point $x$ and two vectors $u$, $v$ in the tangent space at $x$, we can define a metric $G$ on the feature space as follows $ G(x)(u,v) = u^T J(x)^T J(x) v$, where $J(x)$ is the Jacobian of the network at $x$. This metric is known as the pullback metric induced by the network.\\

The pullback metric allows us to measure lengths and angles in the feature space, which provides a way to quantify the geometric transformations applied by the network. In particular, the eigenvalues and eigenvectors of the metric tensor provide information about the local scaling and rotation of the data manifold. This can be used to analyze the expressivity and inductive bias of the network. Furthermore, the geodesics of the pullback metric, which are the shortest paths between points in the feature space, provide a way to interpolate and generate new data points. This is related to the concept of adversarial examples in deep learning: small changes in the input space can lead to large changes in the output space, along the directions of high curvature of the data manifold.\\

\begin{figure*}[ht]
\centering
\includegraphics[width=1\textwidth]{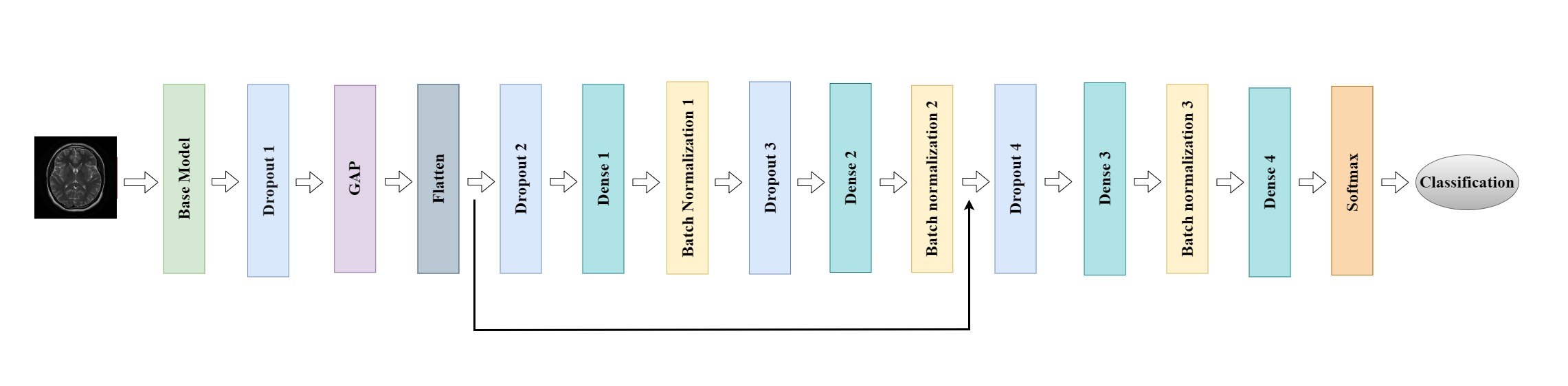}
\caption{Schematic representation of proposed methodology } \label{fig:fig3}
\end{figure*}

In the context of DeepMediX model as shown in figure \ref{fig:fig3}, the convolution layers are building representations of local regions of the data manifold (through filters that detect local features such as edges or textures in images), while the dense layers towards the end of the model are learning global representations (relationships between the features detected by the convolution layers). Given a finite dataset, $\{x^{(i)}\}_{i=1}^m$ with $x^{(i)} \in \mathbb{R}^n$, we consider that each data point lies in a lower-dimensional manifold $\mathcal{M}$ embedded in the high-dimensional input space $\mathbb{R}^n$. Every data point can be depicted by a group of coordinates $(\theta_1, \theta_2, \ldots, \theta_d)$ within a local coordinate framework on the manifold, where $d$, significantly smaller than $n$, is the inherent dimension of the manifold. This local coordinate representation is generally more meaningful and easier to work with than the original high-dimensional representation.\\

We can now consider the base model and the additional layers in our model as functions that operate on the manifold coordinates of the data. The  base model $f_{\text{base}}: \mathcal{M} \rightarrow \mathcal{N}$ transforms the data from the original manifold $\mathcal{M}$ to another manifold $\mathcal{N}$ in the space of feature vectors produced by the base model. This can be considered a form of manifold learning, where the base model learns to map the complex structure of the data manifold $\mathcal{M}$ to a potentially simpler structure in the feature space. The additional layers $f_{\text{add}}: \mathcal{N} \rightarrow \mathcal{P}$ further transform the feature vectors from manifold $\mathcal{N}$ to another manifold $\mathcal{P}$ in the output space. This can be considered as learning a mapping from the feature manifold to the manifold of output classes. Therefore, the entire DeepMediX model can be represented as a composition of these functions $f_{\text{model}} = f_{\text{add}} \circ f_{\text{base}}$, which maps the data from the original manifold $\mathcal{M}$ to the output manifold $\mathcal{P}$. Within the scope of federated learning, each client develops a local model based on its own segment of the data manifold. These individual models are subsequently compiled on the server to construct a comprehensive global model that represents the whole data manifold.

\section{Experiments and results}
We conducted experiments and kept track of the best performing model in terms of validation accuracy using Checkpoints. The epochwise validation accuracy was monitored and training was stopped if the accuracy didn't improve for 15 epochs. Also, the learning rate was reduced by a factor of 3 every 3 epochs if validation accuracy didn't improve. This ensured that the resources weren't wasted following the wrong hyperparameters. The model was run for 50 epochs on kaggle using 2 T4 GPU's.

\subsection{Datasets}
In order to ensure the robustness of our model, we employed a variety of benchmark datasets, including ISIC2018, ISIC2019, HAM10000, and several open-source brain MRI image datasets procured from Kaggle and others \cite{b30}. These datasets collectively offer a comprehensive spectrum of diverse data points and enable an in-depth evaluation of the model's performance across different domains. The International Skin Imaging Collaboration (ISIC) datasets of 2018 and 2019 are prominent resources for melanoma detection and skin lesion analysis, sourced from leading clinical centers worldwide. The ISIC2018 dataset includes images of various benign and malignant skin lesions such as Melanoma, Melanocytic nevus, Basal cell carcinoma, among others, paired with diagnostic labels and metadata like age, sex, and lesion location. The ISIC2019 dataset expands on its predecessor, offering more images and additional diagnostic categories, including Squamous cell carcinoma. The HAM10000 dataset, another key resource, offers a balanced set of 10,015 dermatoscopic images labeled with one of seven disease classes, framed for binary classification: malignant (Melanoma) or benign (all else). Across all these datasets, the evaluation metrics we primarily focused on were accuracy, sensitivity, specificity, and area under the ROC curve. The Br35H \cite{b30} dataset was designed as part of a Brain Detection Challenge in 2020. This is a standard dataset used for binary classification of Brain Tumor MRI Images.

\subsection{Performance}
Our custom deep learning model leveraging MobileNetV2 as the base model has demonstrated superior performance across a range of medical imaging datasets as shown in \ref{tab:tab1}. The structure and complexity of our network design have allowed for high precision, recall, and F1-score performance. In particular, our network has shown extraordinary performance in both the binary and multi-class classifications of Brain tumor as shown in tables  \ref{tab:tab2} and \ref{tab:tab3}, along with Skin Cancer image datasets.

\begin{table}[H] 
	\centering
  \caption{Quantitative Results analysis of the proposed model}
		\label{tab:tab1}
		\resizebox{.45\textwidth}{!}{
        \begin{tabular}{ p{3cm}|p{1.9cm}|p{1.4cm}|p{1.4cm}|p{1.4cm}|p{1.9cm} }
\hline
\textbf{Method}& \textbf{Precision}&\textbf{Recall}&\textbf{F1 Score} &\textbf{ROC-AUC}& \textbf{Accuracy} \\
\hline
ISIC-2019 \cite{b32} & 66.05&63.58&64.58&79.86&79.08 \\
\hline
HAM-10000 \cite{b31}& 90.75&90.66&90.672 &90.66& 90.68 \\
\hline
Brain \cite{b33} &99.01&98.96&98.98 &99.36&99.04 \\
\hline
Br35H \cite{b30} & 99.34&99.34&99.34 &99.34&99.34 \\
\hline
\end{tabular}
	}
\end{table}

 \begin{table}[H] 
	\centering
  \caption{Results in comparison with state-of-the-art (SOTA) using brain (Multi-class) dataset}
		\label{tab:tab2}
		\resizebox{.45\textwidth}{!}{
        \begin{tabular}{ p{3cm}|p{5cm}|p{2cm}}
\hline
\textbf{SOTA}& \textbf{ Model}&\textbf{Accuracy} \\
\hline
Method 1 \cite{b34}  & VGG19 with significant data enhancement  & 94.58 \\
\hline
Method 2  \cite{b35} & Using a genetic algorithm, CNN   & 94.20 \\
\hline
Method 3  \cite{b36} & VGG19 with alterations &94.82 \\
\hline
Method 4  \cite{b37} & Transfer learning via GoogleNet & 97.10 \\
\hline
Method 5  \cite{b38} & Modelling BTC-fCNN with re-training & 98.86 \\
\hline
\textbf{Proposed} & \textbf{DeepMediX} & \textbf{99.04} \\
\hline

\end{tabular}
	}
\end{table}

\begin{table}[H] 
	\centering
  \caption{Results Analysis with other methods \cite{b40} using brain (Br35H: Binary class) dataset .}
		\label{tab:tab3}
		\resizebox{.45\textwidth}{!}{
        \begin{tabular}{ p{2cm}|p{5cm}|p{2cm} }
\hline
\textbf{Method}& Model& \textbf{Accuracy} \\
\hline
  Random&MobileNetV2 + PFpM  & 95.44\\
\hline
 Method 2&Inception-V3 + PFpM &91.44\\
\hline
Method 3 &DenseNet201 + PFpM &92.11\\
\hline
 Method 4 & BMRI-Net model & 99.00 \\
\hline
Method 5 & ResNet50+ PFpM & 93.56\\
\hline
Method 6 & VGG19+ PFpM&92.87  \\
\hline
\textbf{Proposed}  & \textbf{DeepMediX} & \textbf{99.34} \\
\hline
\end{tabular}
	}
\end{table}

\subsection{Ablation Study}

In order to understand the impact of each component in our network structure, an ablation study was conducted. In this study, we systematically removed or changed one component at a time from our network structure and observed the effect on the model's performance. This method allows us to examine the contribution of each component towards the overall performance of the model. It was observed that the removal of any component, such as the dropout layers, global average pooling, or the batch normalization layers, consistently resulted in a decrease in performance metrics. This implies that each component of our model plays a significant role in contributing to its overall performance. The results of the ablation study underscore the necessity of a balance between complexity and efficiency in the design of a deep learning model.\\

In the ablation table \ref{tab:tab4}, Model-1 is DeepMedix utilising Global Max Pooling instead of Global Average Pooling after dropout 1, Model-2 is DeepMedix without the skip connection from Flatten to Dropout 4, Model-3 is DeepMedix without the  dropout modules whereas Model-4 is the proposed metholodology without skip connection as well as the dropout modules, and  Model-5 is a variation of DeepMedix replacing each of the Dense layers with a Dense layer with  256 neurons. Model-6 and 7 are further variations of Model-5 having no skip connection and no dropout modules respectively. Model-8 is a variation of Model-5 having no skip connection and no dropout modules. Removing dropout modules here means removing both Batch Normalization and the Dropout layer.
\begin{table}[H] 
	\centering
  \caption{Quantitative Results}
		\label{tab:tab4}
		\resizebox{.5\textwidth}{!}{
        \begin{tabular}{ p{5cm}|p{2cm}|p{1cm}|p{1cm}|p{1cm}|p{2cm} }
\hline
\textbf{Method}& \textbf{Precision}&\textbf{Recall}&\textbf{F1 Score} &\textbf{ROC-AUC}& \textbf{Accuracy} \\
\hline
Model-1 on ISIC-2019 \cite{b32} & 66&63.58&64.58&79.86&79.20 \\
\hline
Model-1 on HAM-10000 \cite{b31}& 89.77&89.76&89.76&89.76& 89.76 \\
\hline
Model-1 on Brain \cite{b33} &98.86&98.88&98.87&99.26&98.95 \\
\hline
Model-1 on Br35H \cite{30} & 99.17&99.17&99.17 &99.17&99.17 \\
\hline
Model-2 on Brain \cite{b33} & 98.80&98.76&98.78&99.19&98.86 \\
\hline
Model-3 on Brain\cite{b33} & 98.60&98.52&98.56 &99.03&98.67 \\
\hline
Model-4 on Brain\cite{b33} & 98.57&98.43&98.49 &98.97&98.56 \\
\hline
Model-5 on Brain\cite{b33} & 98.87&98.87&98.87 &99.25&98.93 \\
\hline
Model-6 on Brain\cite{b33} & 98.28&98.19&98.23 &98.97&98.33 \\
\hline
Model-7 on Brain\cite{b33} & 98.86&98.77&98.81 &99.19&98.86 \\
\hline
Model-8 on Brain\cite{b33} & 98.49&98.46&98.47 &98.99&98.56 \\
\hline
\end{tabular}
}
\end{table}

\subsection{Computational Efficiency}

Given the goal of designing a model that is both accurate and computationally efficient, we also evaluated the computational demands of our model. The model demonstrated fast inference times, and its small memory footprint makes it suitable for deployment on handheld devices. This opens the door for real-time diagnostics, bringing sophisticated imaging analysis to point-of-care settings. The peak number of FLOPs required by the model was 0.613 Giga FLOPs. The training took 200ms per batch (each batch comprised of 32 images) on kaggle notebook using 2 T4 GPU's.

\section{Conclusion and future work}

This study was centered around the construction and evaluation of a deep learning model, DeepMediX, intended for medical image classification. The performance of the model was tested on brain MRI and skin cancer image datasets, demonstrating promising results. The model not only exhibited strong performance on binary classification tasks using both types of datasets but also maintained its predictive accuracy while remaining computationally efficient. This combination of high accuracy and computational efficiency distinguishes our model within the field of medical image classification, especially valuable in resource-limited settings. In addition to the binary tasks, our model proved proficient in handling multiclass classification problems using skin cancer datasets. This versatility indicates that DeepMediX can be adjusted to cope with more complex classification scenarios. Incorporating federated learning into our model broadens its practical relevance and prospective influence. With federated learning, the DeepMediX model can be trained on a vast amount of distributed data, possibly improving its performance without compromising data privacy. The suitability of DeepMediX for deployment on handheld devices enables the possibility of bringing sophisticated image analysis tools closer to point-of-care settings. Its adaptability makes it an excellent candidate for integration into mobile health (mHealth) applications, thereby broadening the access to advanced diagnostic support. \\

However, our results should be interpreted considering the study's limitations. Future work should focus on testing the performance of the model with a broader array of medical imaging datasets, extending beyond just brain MRIs and skin cancer images. The impacts of architectural adjustments and exploring different federated learning methods on the model's performance also warrant investigation. Moreover, further study is needed to address the unique challenges brought about by federated learning, such as communication efficiency, data heterogeneity, and privacy and security issues. Additionally, the current model's understanding and learning of the data manifold in the high-dimensional space needs to be investigated more deeply to fine-tune the model's ability to understand complex data structures.


\section*{Acknowledgements}
This research has been funded in part by the Ministry of Education, India, under grant reference number  OH-31-24-200-428 and the Department of Atomic Energy, India, under grant number 0204/18/2022/R\&D-II/13979.



\end{document}